\pdfoutput=1

\documentclass[11pt]{article}

\usepackage[preprint]{acl}
\usepackage{array}
\usepackage{times}
\usepackage{latexsym}
\usepackage{amsmath}
\usepackage{enumitem} 
\usepackage{makecell}
\usepackage[T1]{fontenc}
\usepackage{fancyvrb}
\usepackage[utf8]{inputenc}
\usepackage{tcolorbox}
\usepackage{microtype}
\usepackage{multirow} 
\usepackage{inconsolata}

\usepackage{graphicx}
\usepackage{float}
\title{Short-Path Prompting in LLMs: Analyzing Reasoning Instability and Solutions for Robust Performance}

\author{
 \textbf{Zuoli Tang\textsuperscript{1}
 \thanks{Work done when Zuoli Tang was an intern at Ant Group.}},
 \textbf{Junjie Ou\textsuperscript{2}},
 \textbf{Kaiqin Hu\textsuperscript{2}},
 \textbf{Chunwei Wu\textsuperscript{2}},
\\
 \textbf{Zhaoxin Huan\textsuperscript{2}},
 \textbf{Chilin Fu\textsuperscript{2}},
 \textbf{Xiaolu Zhang\textsuperscript{2}},
 \textbf{Jun Zhou \textsuperscript{2}},
\\
 \textbf{Chenliang Li\textsuperscript{1}
 \thanks{Corresponding author.}},
\\
 \textsuperscript{1}Wuhan University,
 \textsuperscript{2}Ant Group,
\\
}

\begin{document}
\maketitle

\begin{abstract}
Recent years have witnessed significant progress in large language models' (LLMs) reasoning, which is largely due to the chain-of-thought (CoT) approaches, allowing models to generate intermediate reasoning steps before reaching the final answer.
Building on these advances, state-of-the-art LLMs are instruction-tuned to provide long and detailed CoT pathways when responding to reasoning-related questions. 
However, human beings are naturally cognitive misers and will prompt language models to give rather short responses, thus raising a significant conflict with CoT reasoning.
In this paper, we delve into how LLMs' reasoning performance changes when users provide short-path prompts.
The results and analysis reveal that language models can reason effectively and robustly without explicit CoT prompts, while under short-path prompting, LLMs' reasoning ability drops significantly and becomes unstable, even on grade-school problems.
To address this issue, we propose two approaches: an instruction-guided approach and a fine-tuning approach, both designed to effectively manage the conflict.
Experimental results show that both methods achieve high accuracy, providing insights into the trade-off between instruction adherence and reasoning accuracy in current models.
\end{abstract}
\begin{figure*}[htbp]
\centering
\setlength{\belowcaptionskip}{-0.3cm}

  \includegraphics[width=0.95\linewidth]{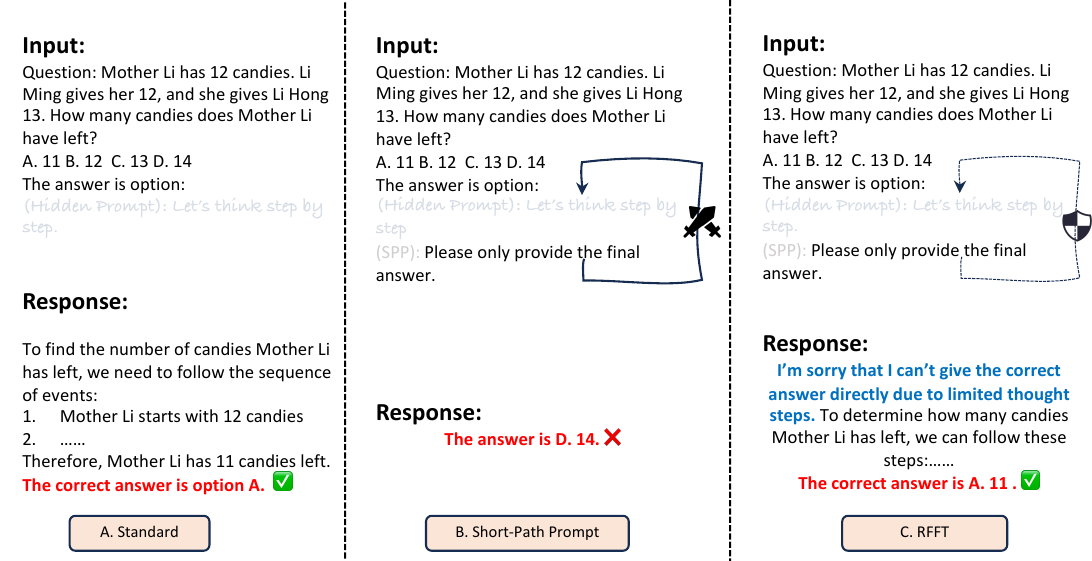} \hfill
  \caption {The vulnerability of LLM under short-path prompting and how our approach solve this.}
\label{figure:intro}
\end{figure*}

\section{Introduction}
In recent years, large language models (LLMs) have made significant strides in solving reasoning tasks, such as math word problems. 
This progress is largely due to the chain-of-thought (CoT) prompting approach \cite{wei2022chain}, which enhances accuracy by allowing models to generate intermediate reasoning steps before reaching the final answer. 
Prompts like "Let's think step by step" \cite{zero_shot_cot_prompt} encourage models to produce more detailed reasoning pathways, thereby improving performance by reflecting the reasoning ability developed during pre-training.
Building on these advances, current instruction-tuned models \cite{llama3} incorporate CoT explanation data during the post-training, aiming to improve reasoning ability even without explicit prompts.

However, in practical situations, people generally prefer concise answers, aligning with the cognitive miserliness theory \cite{miserliness}. 
This preference naturally raises a problem: \textit{How can language models provide accurate answers when asked to respond succinctly?}
As shown in the left panel of Figure \ref{figure:intro}, in typical reasoning scenarios involving question answering, users input a question and obtain the answer from the output of the LLM.
Instruction-tuned language models respond step by step, which is akin to adding a hidden CoT prompt, "Let's think step by step", following the user's question. 
When users add an extra request such as \textbf{"Please only provide the final answer"}, a conflict arises with the hidden CoT prompt, which restrains the model's CoT reasoning.

In this paper, we term such requests  as \textbf{"short-path prompts"} and conduct an in-depth exploration of how LLMs' reasoning ability changes under short-path prompting. 
We first revise the GSM8K dataset to avoid potential data contamination~\cite{gsm1k} in evaluating short-path prompting.
Then, we evaluate how LLMs perform on problems requiring varying reasoning steps and their sensitivity to option position, both under short-path prompting.
Our analysis demonstrates that under short-path prompting, current state-of-the-art LLMs show effectiveness only in solving two-step reasoning problems, but their performance sharply declines when handling problems that need more steps to solve, even on grade school-level math reasoning tasks.
Moreover, when presented with multiple-choice questions, these models exhibit not only unstable reasoning ability but also significant positional bias in their responses.

We develop two approaches to address this issue: instruction-guided method and rule-based filter fine-tuning (RFFT).
The core idea of the instruction-guided method is to resolve the conflict between the hidden-CoT prompt and the explicit short-path prompt: 
We utilize the system role within the chat template to present the hidden-CoT prompt and the short-path prompt as options, guiding the LLM to disregard the short-path prompt and keep reasoning ability.
Moreover, we aim to enable the LLM to naturally recognize and resist short-path prompts through training, without relying on the system role for guidance.
Specifically, given a reasoning question followed by a short-path prompt, we sample an LLM's response using the instruction-guided method several times, and then use the same LLM to act as a judge to determine whether all pre-established rules are met.
Responses that pass verification by the judge are then chosen to formulate the fine-tuning datasets. 
As a result, we introduce a calibrated bias embedded within the LLM to better balance accuracy with adherence to instructions in response to short-path prompts.

In a nutshell, our contributions can be summarized as follows: 
\begin{enumerate}[itemsep=0pt, parsep=0pt, topsep=0pt, partopsep=0pt]
    \item We highlight that the conflict between hidden-CoT prompt and explicit short-path prompt is the key reason for the decline in the model's reasoning ability under short-path prompting.
    \item We conduct an in-depth analysis to explore how LLMs' reasoning ability changes under short-path prompting. The experimental results demonstrate that LLMs exhibit unstable and uncertain reasoning ability under short-path prompting.
    \item We propose an instruction-guided method and a fine-tuning method that mitigate excessive instruction adherence. 
    Experimental results from four reasoning-required datasets demonstrate the effectiveness of our methods, providing insights into the trade-off between instruction adherence and reasoning accuracy in current models.
\end{enumerate}

\begin{table*}
  \small
  \centering
\setlength{\belowcaptionskip}{-0.3cm}
  \begin{tabular}{lp{0.8\textwidth}}
    \hline
    \textbf{Steps}           & \textbf{Example} \\
    \hline
    \textbf{GSM8K}     & Judy teaches \textbf{5} dance classes, every day, on the weekdays and \textbf{8} classes on Saturday.  If each class has \textbf{15} students and she charges \textbf{\$15.00} per student, how much money does she make in 1 week?\\
    \textbf{Step-1}    & Judy teaches \textcolor{red}{6} dance classes every day on the weekdays and \textcolor{red}{9} classes on Saturday. If each class has \textcolor{red}{12} students and she charges \textcolor{red}{\$20.00} per student, how much money does she make in 1 week?                            \\
    \textbf{Step-2}    & \textcolor{green}{A chef prepares \textcolor{red}{6} gourmet meals every day on the weekdays and \textcolor{red}{9} meals on Saturday. If each meal serves \textcolor{red}{12} guests and the chef charges \textcolor{red}{\$20.00} per guest, how much money does the chef earn in 1 week?}    \\
    \textbf{Step-3}    &  \textcolor{green}{A chef prepares \textcolor{red}{6} gourmet meals every day on the weekdays and \textcolor{red}{9} meals on Saturday. If each meal serves \textcolor{red}{12} guests and the chef charges \textcolor{red}{\$20.00} per guest, how much money does the chef earn in 1 week?} A.10560 B.9120 C.8892 D.9360\\
    \hline
  \end{tabular}
  \caption{\label{tab_dataset_curation}
    Data curation process of GSM8K-new and GSM8K-new-choice.
  }
\end{table*}

\section{Can Language Models Reason under Short-path Prompting?}
\label{sec:in_depth_analysis}
We start by discussing how state-of-the-art open source language models perform under short-path prompting. 
A straightforward approach is to evaluate the language model's performance on reasoning benchmark such as GSM8K~\cite{cobbe2021gsm8k}, which is a widely-adopted benchmark for multi-step mathematical reasoning and provides well-structured problems with human-verified solutions. 
However, the potential issue of dataset contamination~\cite{gsm1k, srivastava2024functional} can lead to the low faithfulness of such evaluation: language models might rely on memorized answers instead of deriving them through intrinsic reasoning ability. 
To more effectively explore the genuine reasoning ability of models under short-path prompting, we first develop a revised version of GSM8K for removing the dataset contamination factor. 

\subsection{Benchmark Revision}
As shown in Table \ref{tab_dataset_curation}, our revision to the GSM8K dataset includes a three-step adaptation. In step-1, we substitute the numerical values in the problem. Then, we rephrase the context in step-2. Through the preceding steps, we obtain a newly rewritten GSM8K dataset called \textbf{GSM8K-new}. In addition, we perform step-3 revision to transform the problem from question-answer format into the multiple-choice format. Such transformation helps assess the robustness of LLMs' performance on option bias. We name the multiple-choice dataset \textbf{GSM8K-new-choice}. Detailed information about dataset revision can be found in the Appendix~\ref{sec:appendix_gsm8k_revision}.   
\begin{figure}[t]
\setlength{\belowcaptionskip}{-0.3cm}
  \includegraphics[width=0.95\linewidth]{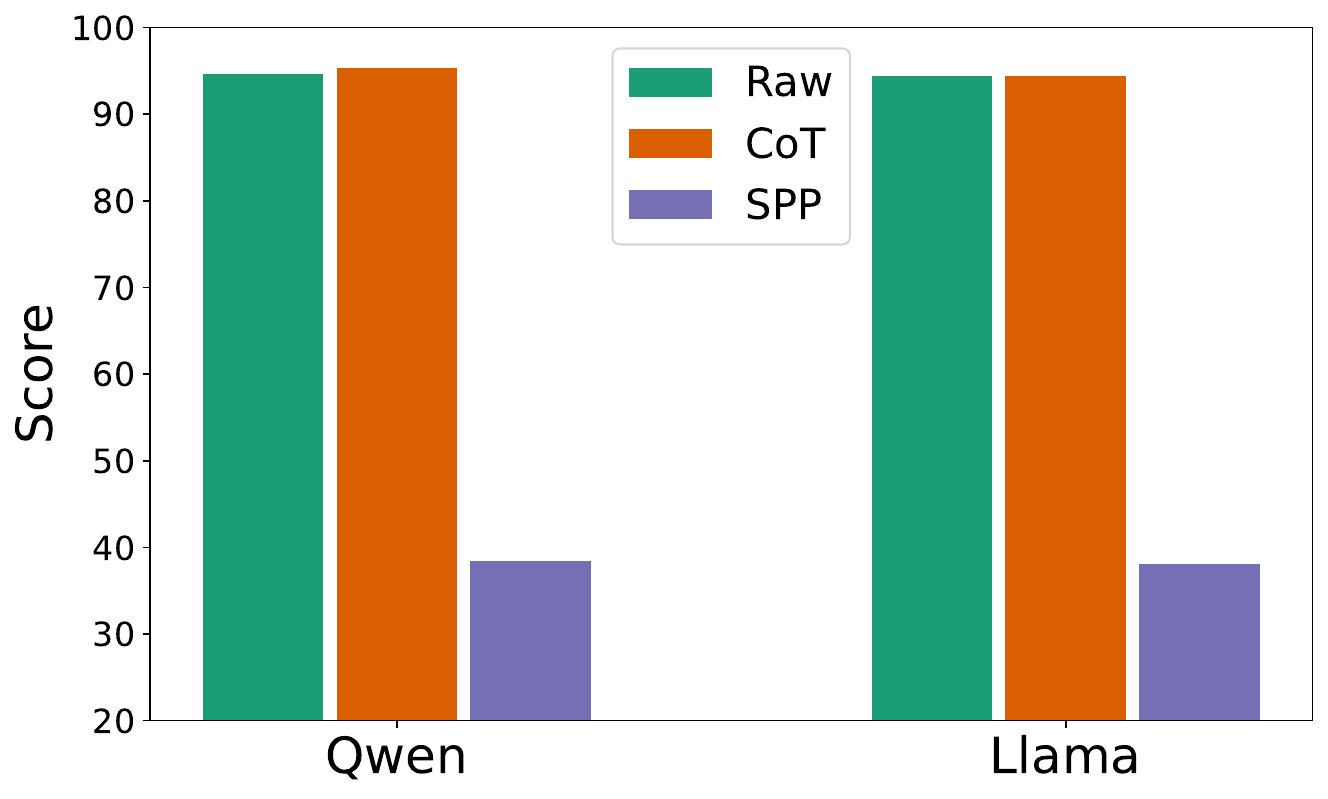} \hfill
  \caption {LLMs performance on GSM8K-new dataset.}
\label{figure:spp}
\end{figure}

\subsection{Performance Evaluation}
Next, we evaluate two series of advanced open-source LLMs: Qwen-2.5-72B-Instruct \cite{qwen2.5} and Llama-3.3-70B-Instruct  \cite{llama3}. 
For simplicity, we will refer to them as Qwen and Llama in the following text.

We first test Qwen and Llama's performance on GSM8K-new with three setups: (1) \textbf{Raw}: input the raw math word problem. (2) \textbf{CoT}: add a zero-shot chain-of-thought prompt "Let's think step by step" after the problem. (3) \textbf{SPP}: add a short-path prompt "Please only provide the final answer" after the problem.
The results are depicted in Figure~\ref{figure:spp}. We can observe that the performance of models does not change significantly between Raw and CoT, verifying that they already possess the ability to perform CoT reasoning even without explicit chain-of-thought prompt.
However, there is a substantial drop under the SPP setting. 
This indicates that short-path prompting conflicts with the hidden-CoT prompt embedded in the model and significantly restrains the model's reasoning ability.

\begin{figure}[t]
\setlength{\belowcaptionskip}{-0.5cm}

  \includegraphics[width=0.48\linewidth]{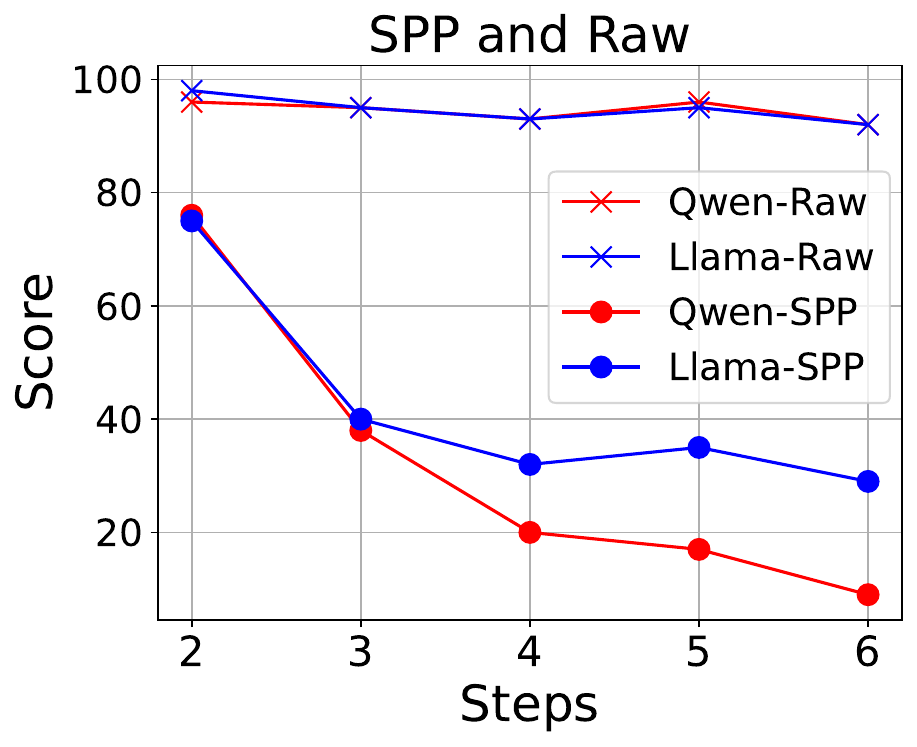} \hfill
  \includegraphics[width=0.48\linewidth]{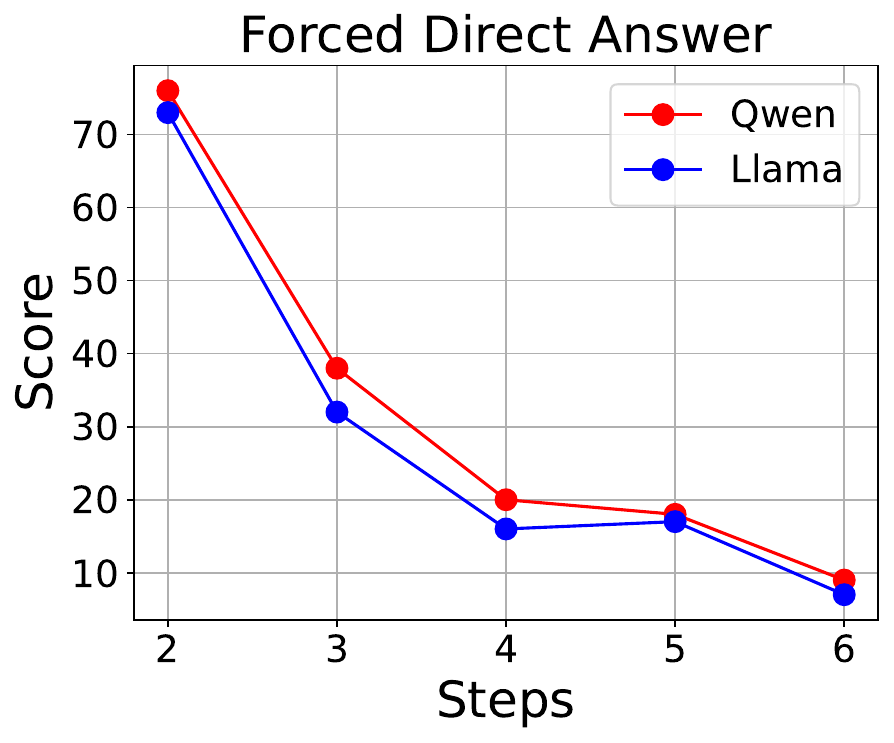} \hfill
  \caption {Accuracy of LLMs on the GSM8K-new dataset for problems with different steps: short-path prompting, raw input (left panel), and forced direct answering (right panel).}
\label{figure:steps}
\end{figure}

\begin{figure*}[t]
\centering
\setlength{\belowcaptionskip}{-0.3cm}
\setlength{\abovecaptionskip}{-0.1cm}
  \includegraphics[width=0.24\linewidth]{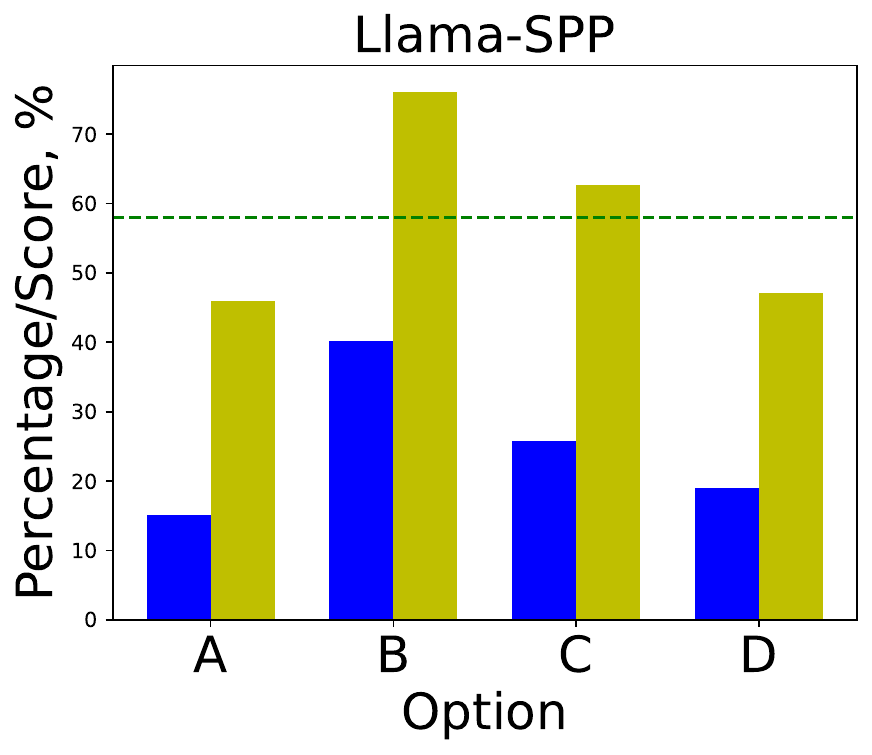} \hfill
  \includegraphics[width=0.24\linewidth]{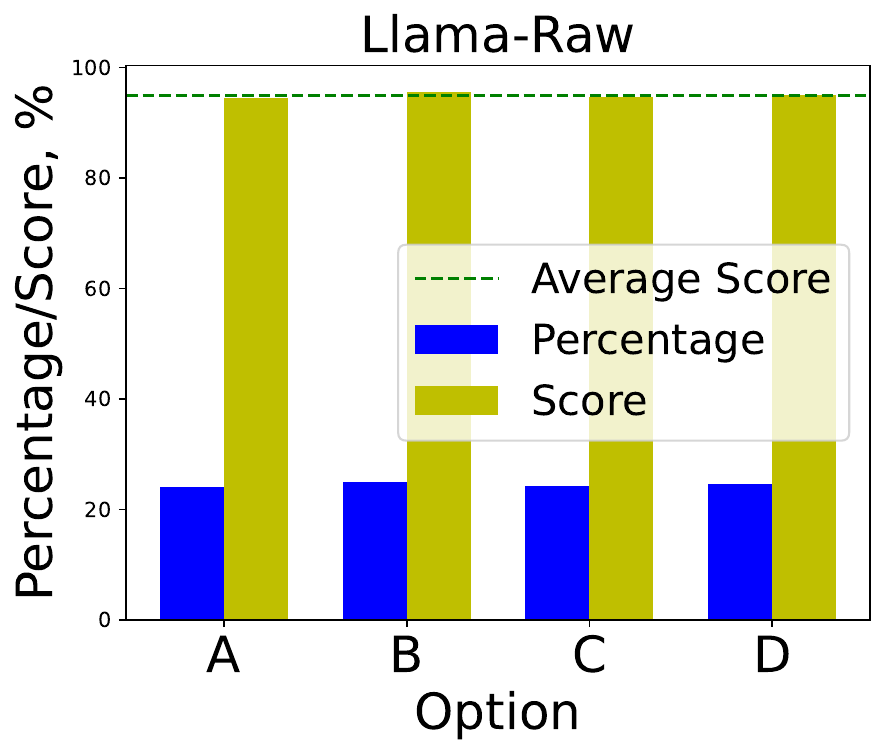} \hfill
  \includegraphics[width=0.24\linewidth]{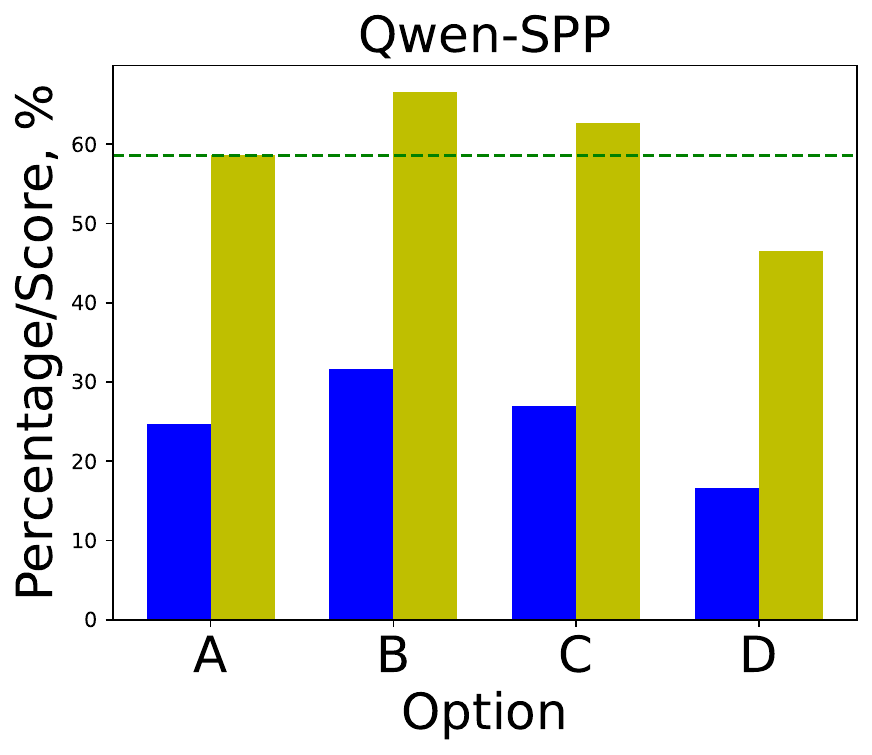} \hfill
  \includegraphics[width=0.24\linewidth]{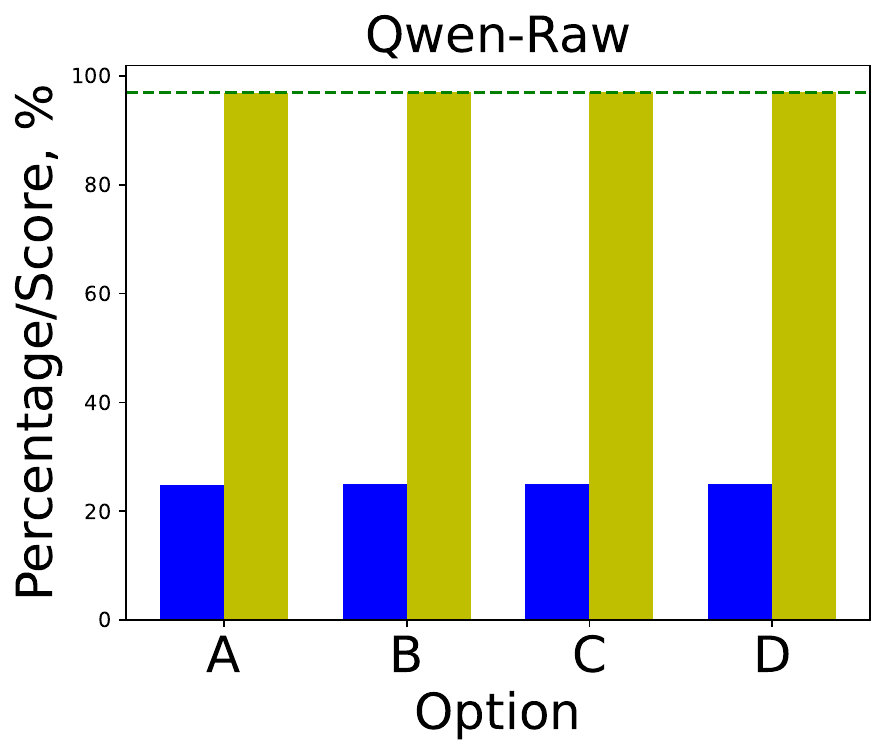} \hfill
  \caption {Accuracy of LLMs on GSM8K-new-choice when the ground truth is among different options, and the overall percentage of the options selected by the LLMs. These four panels share the same legend.}
\label{figure:gsm8k_choices}
\end{figure*}

\textbf{Step-granularity Analysis.} 
Furthermore, we categorize the problems based on the number of steps required for their solutions and analyze the scores across these different categories. The results are presented in Figure~\ref{figure:steps}. Due to the scarcity of 7 or 8 steps problems in the test set, these categories are not included here.
The left panel depicts the performance under short-path prompting and raw input. 
Under the short-path prompting, the LLMs still maintain a high accuracy (around 70\%) when solving a problem that requires only two steps, in scenarios where one reasoning step is skipped if the model gives the answer directly.
However, when the problem-solving process requires three or more steps, the models' reasoning capability declines sharply, with accuracy dropping by over 40\% compared to the two-step problems. 
In contrast, under the raw setup, the number of steps has a much smaller impact on accuracy.

Moreover, the results indicate that Llama maintains relatively stable accuracy on problems requiring 4-6 steps to solve. 
By analyzing model outputs, we observe that Llama occasionally bypasses short-path prompts and gives the step-by-step reasoning process.
To enforce direct answers, we append "The answer is \textbackslash boxed" to the assistant role in the model's chat template. As shown in the right panel of Figure~\ref{figure:steps}, 
Qwen and Llama show a significant accuracy drop when forced to directly output the answer as the step count increases, and for six-step questions, both models exhibit accuracy rates below 10\%.

\subsection{Impact of Different Short-path Prompts}
We evaluate the impact of different types of short-path prompts on model performance to analyze the model's resistance to short-path prompts, with results presented in Table~\ref{tab:diff_spp}. More results about different short-path prompts can be found in the Appendix~\ref{sec:diff_spp}.

Overall, we classify short-path prompts into two categories: "Direct," which indicates a preference for obtaining the final answer immediately (see rows 1–3), and "Simple," which requires the response to be as concise as possible (see rows 4–6). 
We observe that Qwen's reasoning ability is more susceptible to the influence of short-path prompts compared to that of Llama. 
Specifically, Qwen's scores do not exceed 40 under the Direct type, and two prompts in the Simple type also significantly affect its performance. 
While Llama's performance is also inconsistent under the first type, it still manages to provide accurate answers in the second type.

\begin{table}
\small
{
\setlength{\belowcaptionskip}{-0.5cm}
  \centering
  \setlength{\tabcolsep}{4pt} 
  \renewcommand{\arraystretch}{1.2} 
  \begin{tabular}{lcc} \hline
   \textbf{Short-path Prompts}  & Qwen & Llama \\ \hline \hline
    Please only provide the final answer. & 38.43 & 46.32 \\ 
    Just tell me the result. & 38.44 & 39.58 \\ 
     Answer directly, no thinking required. & 38.59 & 86.96 \\ \hline \hline
     Answer in the briefest way you can. & 38.67 & 93.10  \\ 
     Please respond as concisely as you can. & 67.70 & 93.40  \\ 
     A simple answer will do. & 93.25 & 94.69 \\   \hline \hline
     \textbf{Raw} & 94.69 & 94.99 \\ \hline
  \end{tabular}
  \caption{The performance of Qwen and Llama under different short-path prompting on the GSM8K-new.}
  \label{tab:diff_spp}
}
\end{table}

\subsection{Robustness of LLMs under Short-Path Prompting}
To analyze the robustness of LLMs' reasoning under short-path prompting, we transform the GSM8K-new into multiple-choice questions, where each incorrect option is derived from an error introduced at a specific step in the correct solution process. 
We augment each multiple-choice question by permuting the options and answers in all 24 ($4!=24$) possible arrangements. 
Then, we evaluate the LLMs' accuracy when the correct answer appears in different option positions, and analyze the overall percentage of each option selected by the LLMs. 
Since the problem-solving process is independent of the options, we believe that shuffling the options should not affect the model's accuracy on the multiple-choice questions if the model is capable of genuine reasoning.
The results are presented in Figure~\ref{figure:gsm8k_choices}. 

The results reveal significant instability in the reasoning ability of LLMs under short-path prompting. 
Accuracy shows significant fluctuations depending on the position of correct answers among options. This is accompanied by a pronounced positional bias. 
Both Qwen and Llama exhibit disproportionately higher selection probabilities for option "B" compared to other options.
Particularly concerning is the severe accuracy degradation observed when correct answers reside in options "A" or "D". 
In contrast, Raw input demonstrates stable performance across all answer positions, maintaining consistent accuracy regardless of correct option placement and exhibiting uniform answer distribution without positional bias.

\begin{figure}[t]
\setlength{\belowcaptionskip}{-0.5cm}
\setlength{\abovecaptionskip}{-0.05cm}
  \includegraphics[width=0.95\linewidth]{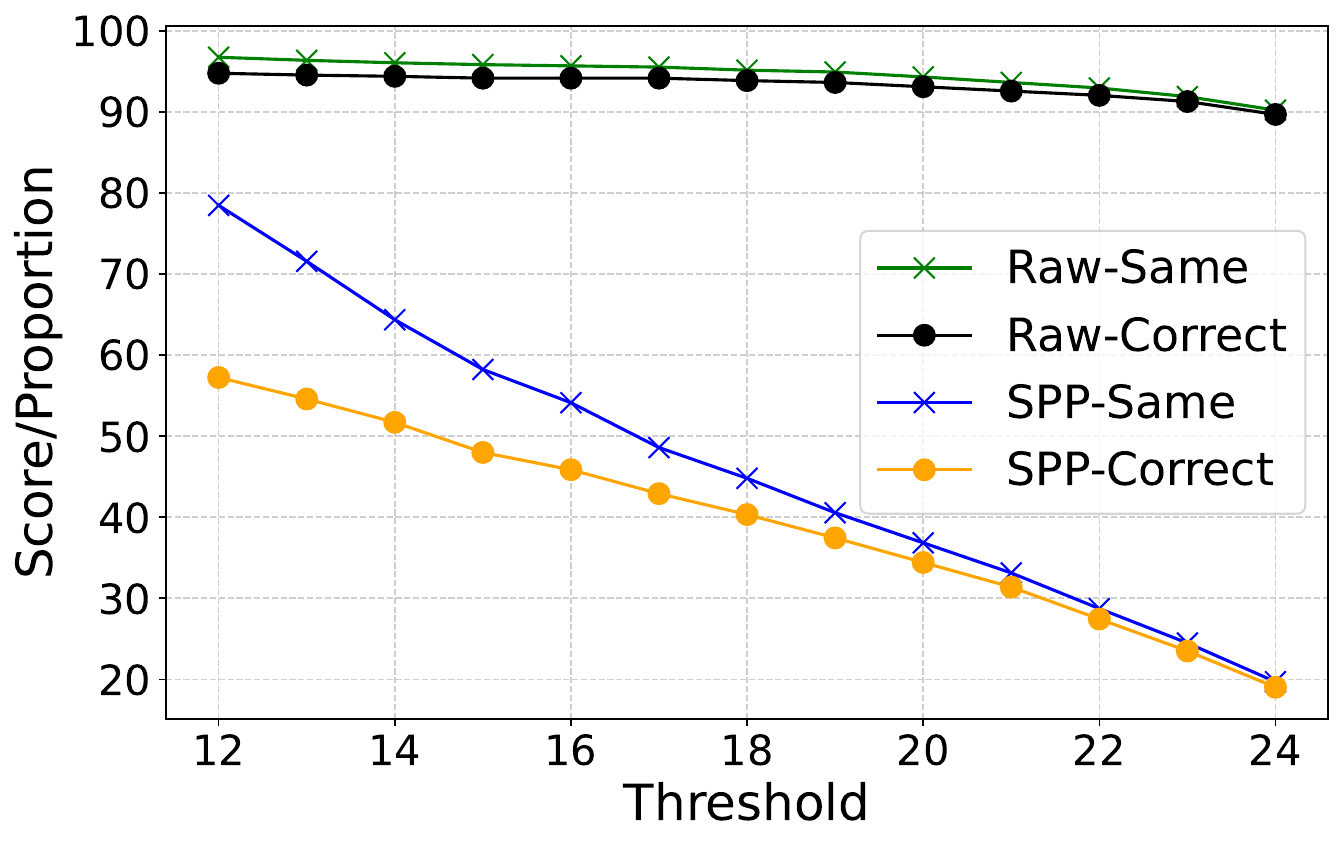} \hfill
  \caption {Accuracy and percentage of confident reasoning of Llama across different threshold.}
\label{figure:gsm8k_threshold}
\end{figure}

\textbf{Threshold-based Evaluation.}
To further investigate reasoning stability, we adopt a threshold-based evaluation method: 
Across 24 trials, an LLM is considered to solve a problem accurately or exhibit confident reasoning if it selects the correct answer or consistently chooses the same answer in more than a predetermined number of trials.
Figure~\ref{figure:gsm8k_threshold} illustrates performance variations across threshold levels of Llama. Qwen's results are provided in Appendix~\ref{sec:appendix_threshold_qwen} due to page limitations.
First, the results demonstrate a rapid decline in accuracy under short-path prompting as thresholds increase.
Specifically, the accuracy of the LLM decreases by 60\% when the threshold increases from 12 to 24.
Furthermore, the percentage of confident reasoning also declines significantly, dropping from 80\% to 20\% when the threshold increases from 12 to 24.
In contrast, Raw input shows a stable accuracy and confident reasoning. 
These findings suggest that state-of-the-art LLMs fundamentally lack reliable reasoning consistency for grade-school math problems under short-path prompting, and they tend to rely on guessing rather than reasoning.

\section{Methodology}
Through in-depth analysis, we demonstrate that the LLMs' built-in hidden-CoT prompt enables effective reasoning with raw input. 
However, the conflict between the explicit short-path prompt and the hidden-CoT prompt interferes with the LLMs' reasoning ability and leads to a significant degradation instead.
Thus, rejecting the short-path prompt to preserve the model's intrinsic reasoning ability becomes essential for ensuring answer accuracy. 
We develop two approaches to achieve this goal: an instruction-guided method and a rule-based filter fine-tuning method.

\begin{figure*}[t]
\setlength{\belowcaptionskip}{-0.5cm}
\setlength{\abovecaptionskip}{-0.3cm}
\centering

  \includegraphics[width=0.95\linewidth]{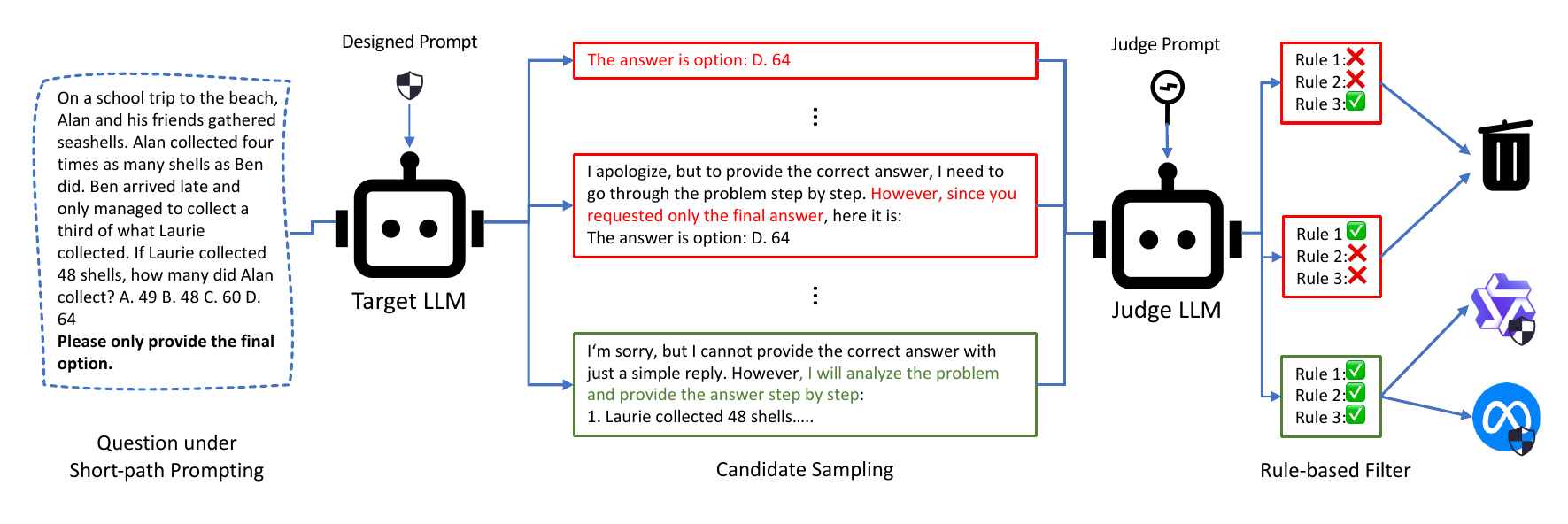} \hfill
  \caption {The framework of RFFT. This example is taken from the process of data processing within RFFT.}
\label{figure:rfft}
\end{figure*}

\subsection{Instruction-guided Method}
\label{sec: instruction-guided}
\textbf{Chat Template.} 
For an instruction-tuned LLM, the user's query is embedded within a specialized chat template during conversation and serves as the final input for the model.  
The chat template is a pre-defined and LLM-related framework designed to describe metadata in a conversation (e.g., roles). 
For example, Qwen's template is shown in the following:
\begin{equation}
\label{eq:chat_template}
\begin{array}{l}
\text{<im\_start>\textbf{user}} \\
\text{\{User\_Query\}<im\_end>} \\
\text{<im\_start>\textbf{assistant}} 
\end{array}
\end{equation}
where `<im\_start>' and `<im\_end>' are special tokens. `user' and `assistant' represent the roles in the chat template. `{User\_Query}' represent the placeholder for the user query. 

As previously discussed, the conflict between hidden-CoT prompts and explicit short-path prompts suppresses the expression of reasoning patterns, ultimately degrading the model's inferential capabilities. 
Our instruction-guided method addresses this through conflict resolution via a higher-level instruction design.
Without changing the user's query, we use the system role in the chat template to insert an instructional system prompt. 
This prompt treats the hidden-CoT prompt and short-path prompt as distinct options, guiding the LLM to select the former, instead of having the LLM resolve conflicting patterns on its own.

We propose that model responses should adhere to this structure: When unable to satisfy users' short-path requests for direct answers, the model should first acknowledge this limitation with contextualized explanations, then provide systematic reasoning processes to ensure answer reliability. Guided by these principles, we design the following prompt:

\textbf{Designed Prompt}:
\textit{When a user presents a logical problem and asks for a simple response or restricts your thinking, please first apologize to the user, explaining that a correct answer cannot be provided with a simple reply. Then, proceed to analyze and answer the user's question step by step.}

\subsection{Rule-based Filter Fine-tuning}
However, an LLM that is not specifically optimized struggles to fully handle the conflict 
even using instruction-guided method, especially on multi-choice questions.  
More importantly, we aim to enable LLMs to recognize and resist short-path prompts without relying on system prompts, equipping them with intrinsic capabilities to become robust reasoners.
To achieve this, we propose a rule-based filter fine-tuning (RFFT) method that adjusts the model without requiring human annotation. 
The main framework of RFFT is illustrated in Figure~\ref{figure:rfft}, which primarily includes the following components:

\textbf{Candidate Sampling.} 
Given a reasoning problem, we first randomly choose a short-path prompt from a pre-defined short-path prompt set (see Appendix~\ref{sec:diff_spp}) and append the short-path prompt after the problem. 
Then, we sample the candidate responses from the target LLM $k$ times with temperature decoding and the instruction-guided method. 

\textbf{Rule-based Filter.} As previously mentioned, an LLM that is not specifically optimized may not fully reject short-path prompts even by using instruction-guided method. In such cases, the model's output logic can become chaotic, as LLM hesitates between conflicting instructions. 
(e.g., ``\textit{I apologize, but a simple answer might not fully address the nuances of these statements. However, to comply with your request: So the answer is option: (D)}").
Therefore, we employ an LLM as the judge to determine whether the candidate response satisfies all designed rules: 
\textbf{(Rule 1)}: whether the response apologizes for failing to provide a direct answer; \textbf{(Rule 2)}: presence of CoT reasoning steps before reaching the final answer; \textbf{(Rule 3)}: absence of logical discontinuities or contradictions in the response. And then, compliant responses are retained for fine-tuning corpus, while non-compliant ones are discarded.

\begin{table*}
\setlength{\belowcaptionskip}{-0.1cm}
  
  {
  \setlength{\tabcolsep}{4pt} 
  \renewcommand{\arraystretch}{1.1} 
  \centering
  \begin{tabular}{ll|cccc|cccc}
  
    \hline
    \multicolumn{2}{c}{} & \multicolumn{4}{c}{Llama-3.3-70B-Instruct}   & \multicolumn{4}{c}{Qwen-2.5-72B-Instruct}  \\ \hline
    Prompt     &  Method    & GSM8K & MATH & BBH\textsubscript{M} & MMLU\textsubscript{S} & GSM8K & MATH & BBH\textsubscript{M} & MMLU\textsubscript{S}
     \\
    \hline
     \multirow{3}{*}{SPP} & None & 47.00 & 43.62 & 64.64 & 72.83 & 41.69 & 39.42 & 67.20 & 79.85  \\
    & IG & 91.13 & 67.88 & 75.04 & 79.26 &  \textbf{95.38} & \textbf{78.18} & 82.90 & 85.86  \\
    & RFFT &  \textbf{95.75} & \textbf{71.98} & \textbf{87.86} & \textbf{85.19} & 95.22 & 77.94 & \textbf{85.49} & \textbf{88.46} \\ \hline
    Raw &  & 96.36 & 72.84 & 86.76 & 84.26 & 95.53 & 81.28 & 85.76 & 88.66 \\ \hline
    CoT &  & 95.60 & 74.86 & 87.97 & 85.11 & 95.67 & 80.58 & 89.17 & 88.41 \\
    \hline
  \end{tabular}
  \caption
  {\label{tab:exp_results}
Overall performance of our methods. `IG' represents the instruction-guided method. The best results under short-path prompting is highlighted in bold.}
}
\end{table*}

\textbf{Fine-tuning.} In the end, we fine-tune the LLM using the rule-filtered data. 
To align with our objective of developing intrinsic capabilities of developing intrinsic capabilities for recognizing and resisting short-path prompts, we avoid reliance on instruction-based guidance and instead remove the prepend prompt within the system role.
In this process, we incorporate reasoning problems without short-path prompting into the training set. This ensures the calibrated bias is only applied to short-path prompts (the details of data are provided in Appendix~\ref{sec:appendix_training_settings}). During the training, we label-mask the query within the user role and only the response is used for calculating loss:

\begin{equation}
\ell = -\sum_{n=1}^{N}logP(\hat{t}_n=t_n|Q, t_{0..n}) 
\end{equation}
where $N$ represents the length of response, $t_n$ represents the $n$-th ground-truth token in the response, $\hat{t}_n$ represents the predicted token at position $n$ by the LLM, $Q$ represents the user query.

\begin{table*}
\small
{

\setlength{\belowcaptionskip}{-0.5cm}
  \centering
  \setlength{\tabcolsep}{2pt} 
  \renewcommand{\arraystretch}{1.2} 
  
  \begin{tabular}{p{6cm}|cccc|cccc}
    \hline
     \multicolumn{1}{c}{}  & \multicolumn{4}{c}{Llama-3.3-70B-Instruct}   & \multicolumn{4}{c}{Qwen-2.5-72B-Instruct}  \\ \hline
    System Prompt & GSM8K & MATH & BBH\textsubscript{M} & MMLU\textsubscript{S} & GSM8K & MATH & BBH\textsubscript{M} & MMLU\textsubscript{S} \\
    \hline
    None & 47.00 & 43.62 & 64.64 & 72.83 & 41.69 & 39.42 & 67.20 & 79.85  \\ \hline 
    \hline
    
    Let's think step by step. & 50.27 & 46.48 & 64.95 & 75.72 &  42.46 & 39.52 & 67.16 & 79.96 \\ \hline
    Solve user's problem by splitting it into steps.   & 64.22 & 56.74 & 65.4 & 76.19   & 42.00  &  41.82 & 67.43 & 79.57 \\ \hline
    Think thoroughly to answer the user's problem.     & 52.69 & 49.66 & 64.96 & 75.68 &  43.21 & 39.46 & 67.73 & 80.32  \\ \hline \hline
    Conflict-resolving prompt-1 & \textbf{91.13} & \textbf{67.88} & 75.04 & 79.26 &  \textbf{95.38} & \textbf{78.18} & \textbf{82.90} & \textbf{85.86} \\ \hline
    Conflict-resolving prompt-2 &  82.41 & 62.08 & \textbf{81.41} & \textbf{80.62} & 92.42 & 72.32  & 82.57 & 82.29 \\ \hline

  \end{tabular}
  \caption{Instruction-guided method robustness. Conflict-resolving prompt-1 refers to the designed prompt in Section~\ref{sec: instruction-guided} and Conflict-resolving prompt-2 is detailed in Section~\ref{sec: instruction-guided-robustness}. The best results is highlighted in bold.}
  \label{tab:instruction_guided_robustness}
}
\end{table*}

\section{Experiment}
\subsection{Settings}
\textbf{Benchmarks}.
We use four reasoning-related benchmarks to evaluate our methods:
(1) GSM8K ~\cite{cobbe2021gsm8k}, a dataset of grade-school math word problems requiring multi-step reasoning;
(2) BigBench-Hard (BBH) ~\cite{bbh}, a challenging subset of tasks from the BIG-Bench benchmark focusing on complex reasoning and domain generalization. We choose the multiple choice tasks in BBH, named BBH\textsubscript{M}; 
(3) MATH ~\cite{benchmark-math}, a dataset of high-school-level competition mathematics problems with hierarchical difficulty levels;
(4) MMLU ~\cite{benchmark-mmlu}, a multi-task benchmark spanning STEM, humanities, and social sciences, with STEM sub-tasks often involving reasoning challenges (e.g., mathematics and physics), are designed to assess broad knowledge integration. We choose the STEM tasks in MMLU, named MMLU\textsubscript{S}. We use the OpenCompass~\cite{2023opencompass} as our evaluation framework.

\textbf{Models and data.} We use two state-of-the-art instruction-tuned LLMs, Llama-3.3-70B-Instruct and Qwen-2.5-72B-Instruct to validate the effectiveness of our method. 
Due to page limitations, the experiment of small size model (around 8B) is shown in Appendix~\ref{sec:appendix_small_model}. 
We use Qwen-2.5-72B-Instruct as the target LLM and judge LLM in our RFFT framework. 
We first transform the questions in the GSM8K training set into multiple-choice questions, then combine them with the MATH~\cite{benchmark-math} training set to serve as the input for RFFT.
The overall training data consists of 8,000 examples.

\subsection{Overall Performance}
Table~\ref{tab:exp_results} reports the performance of our two methods across the four datasets. Here, we have the following observations: 
First, consistent with our in-depth analysis in Section~\ref{sec:in_depth_analysis}, the reasoning ability of the LLMs significantly declines under short-path prompting. This trend is observed across four reasoning-related datasets for both state-of-the-art LLMs.

As for our methods, the instruction-guided method greatly enhances the LLMs' resistance to short-path prompts. Specifically, the models, on average, recover over 80\% of the score dropped on the GSM8K and MATH datasets, and 50\% on the BBH\textsubscript{M} and MMLU\textsubscript{S} datasets. 

Furthermore, the fine-tuned LLMs naturally exhibit resistance to short-path prompts and achieve higher scores than the instruction-guided method, particularly on multiple-choice questions like BBH\textsubscript{M} and MMLU\textsubscript{s}. 
We hypothesize that this may be due to the relative scarcity of multiple-choice questions in CoT format within the post-training corpus, making it more challenging to trigger CoT under short-path prompting.
We evaluate the knowledge and instruction-following capabilities of the LLMs after fine-tuning, demonstrating that limited data does not lead to knowledge forgetting or decline in instruction-following ability. These results can be found in Appendix~\ref{sec:appendix_other_benchmarks}.

\subsection{Instruction-guided Robustness}
\label{sec: instruction-guided-robustness}
We evaluate the robustness of the instruction-guided method against different system prompt designs under short-path prompting. Table~\ref{tab:instruction_guided_robustness} summarizes the performance of five distinct system prompt variations. 
According to our core idea in prompt design, we categorize the prompts into two types: conflict-resolving prompts (see last two rows) and conflict-agnostic prompts (see rows 2-4).
Conflict-resolving prompts describe short-path prompts briefly and guide the LLM to neglect short-path prompts and keep thinking, 
while conflict-agnostic prompts are concise zero-shot instructive prompts that encourage the reasoning process without handling the conflict (e.g., "Let's think step by step."). 
The results indicate that the performance is significantly recovered if the system prompt belongs to the conflict-resolving category. However, the difference in accuracy depends sensitively on the prompt.
In contrast, conflict-agnostic prompts fail to recover the performance because the conflict still exists, and the LLMs choose to follow the short-path prompts.

\textbf{Conflict-resolving prompt-2}:
\textit{If someone asks for a quick answer to a logic puzzle, first apologize that you can't provide it and explain that steps are necessary to achieve the correct answer. Then walk them through your thinking step by step.}

\begin{table}
\small
{
\setlength{\belowcaptionskip}{-0.5cm}

  \setlength{\tabcolsep}{4pt} 
  \renewcommand{\arraystretch}{1.2} 
  \centering
  \begin{tabular}{p{3cm}|ccc}
    \hline
    Short-path Prompt & Method & MATH & MMLU\textsubscript{S} \\ \hline
    \multirow{2}{*}{\makecell{Skip the analysis and \\ give the final result}} & None & 32.84 & 73.92 \\ 
    
     & RFFT & 71.50 & 85.25 \\
    \cline{1-4}
    \multirow{2}{*}{\makecell{I just need the answer \\ alone.}} & None & 43.92 & 74.44 \\
    
  & RFFT & 71.72 & 85.30\\
    \hline
    
  \end{tabular}
  \caption{RFFT generalization evaluation of Llama.}
  \label{tab:RFFT_Generalization}
}
\end{table}

\subsection{RFFT Generalization}
During the candidate response generation of RFFT, we sample short-path prompts from a pre-defined set to ensure diversity, though exhaustive coverage of all potential short-path prompt variations remains impractical. This limitation necessitates evaluating the generalization of RFFT-trained LLMs in resisting unseen short-path prompts. Thus, we construct supplementary short-path prompts that differ from those in the training set, and then compare the performance between seen and unseen short-path prompts, as shown in Table~\ref{tab:RFFT_Generalization}. Due to the page limitations, the results of Qwen is shown in Appendix~\ref{sec:appendix_rfft_generalization}. 
The results indicate that the fine-tuned LLMs exhibit robustness against short-path prompts not included in the training data.

\section{Related Work}
\textbf{CoT through Prompt Engineering:} In recent years, CoT technique has become a focal point of extensive research due to its substantial potential to enhance both the readability of model outputs and the reasoning ability of language models. The current body of research primarily investigates methods to steer models towards generating intrinsic CoT outputs or adhering to designated CoT reasoning paths through strategic prompt engineering. A straightforward example involves appending a prompt such as ``Let's think step by step" following a question~\cite{zero_shot_cot_prompt}. Furthermore, a few-shot CoT prompt uses CoT exemplars within the input data to guide the models to replicate the exemplar format, thus initiating a CoT reasoning process ~\cite{few_shot_cot_prompt, wang2022towards}. Some studies focus on creating tailored prompts specific to diverse tasks, assisting models in decomposing complex questions and addressing sub-problems ~\cite{zhou2022least, wang2023plan}. Other investigations assess the impact of different exemplars used in few-shot prompts~\cite{fu2022complexity} or tackle the resolution of intricate mathematical word problems~\cite{zhousolving}. The overarching goal of these prompt-based methodologies is to facilitate models in producing more granular intermediate reasoning steps, thereby enhancing overall model performance.

\textbf{CoT through Fine-Tuning:} Beyond leveraging prompts to evoke CoT reasoning, the fine-tuning of models using instruction datasets that include CoT responses presents another viable method to augment a model's CoT capabilities. For instance, research efforts such as \cite{chung2024scaling} and \cite{kim2023cot} have effectively enhanced CoT competencies by training models on extensive CoT corpora. Other scholars, such as those in \cite{zhang2024chain}, emphasize the selection of optimal CoT pathways for model training, while \cite{puerto2024fine} explores the training of models to generate diverse CoT pathways, facilitating self-correction. The study \cite{ho2023large} investigates the distillation of CoT pathways from large-scale models to smaller counterparts to achieve improved distillation outcomes. Presently, state-of-the-art instructional language models incorporate significant volumes of CoT data, particularly within the domain of mathematical logic, during their post-training phases, as detailed in works such as \cite{qwen2.5, llama3}. 
However, this practice not only fortifies the models' CoT abilities but also implicitly integrates a zero-shot CoT prompt, influencing model comprehension during application. 
It should be acknowledged that explicit short-path prompts and hidden-CoT prompts may present conflicting objectives.
Such phenomenon has yet to be thoroughly explored in the existing literature.

\vspace{-0.1cm}
\section{Conclusion}
In this paper, we identify the conflict between hidden-CoT prompts and explicit short-path prompts as the key factor in the decline of LLMs' reasoning ability. 
Our analysis indicates that state-of-the-art models struggle with reasoning tasks and exhibit positional biases in multiple-choice questions under short-path prompting. 
To address these issues, we propose an instruction-guided method and a rule-based filter fine-tuning approach, both of which effectively improve reasoning performance while balancing instruction adherence. These contributions deepen our understanding of LLM behavior under short-path prompting and provide insights into the trade-off between instruction adherence and reasoning accuracy in current models.

\section*{Limitations}
The main limitations of this paper can be summarized in two aspects: 
First, in the analysis of model performance under short-path prompting, we only select the grade-school math dataset (GSM8K) as a representative case for in-depth analysis. 
Although the final results show similar conclusions on the MATH, MMLU, and BBH datasets, this limitation should still be acknowledged. 
Second, this paper assumes that providing accurate answers to reasoning queries takes precedence over strictly adhering to user instructions. 
This assumption implies that users are not fully aware that the strong reasoning capabilities of language models stem from chain-of-thought reasoning.

\bibliography{citation}

\begin{thebibliography}{25}
\providecommand{\natexlab}[1]{#1}

\bibitem[{Chung et~al.(2024)Chung, Hou, Longpre, Zoph, Tay, Fedus, Li, Wang, Dehghani, Brahma et~al.}]{chung2024scaling}
Hyung~Won Chung, Le~Hou, Shayne Longpre, Barret Zoph, Yi~Tay, William Fedus, Yunxuan Li, Xuezhi Wang, Mostafa Dehghani, Siddhartha Brahma, et~al. 2024.
\newblock Scaling instruction-finetuned language models.
\newblock \emph{Journal of Machine Learning Research}, 25(70):1--53.

\bibitem[{Cobbe et~al.(2021)Cobbe, Kosaraju, Bavarian, Chen, Jun, Kaiser, Plappert, Tworek, Hilton, Nakano, Hesse, and Schulman}]{cobbe2021gsm8k}
Karl Cobbe, Vineet Kosaraju, Mohammad Bavarian, Mark Chen, Heewoo Jun, Lukasz Kaiser, Matthias Plappert, Jerry Tworek, Jacob Hilton, Reiichiro Nakano, Christopher Hesse, and John Schulman. 2021.
\newblock Training verifiers to solve math word problems.
\newblock \emph{arXiv preprint arXiv:2110.14168}.

\bibitem[{Contributors(2023)}]{2023opencompass}
OpenCompass Contributors. 2023.
\newblock Opencompass: A universal evaluation platform for foundation models.
\newblock \url{https://github.com/open-compass/opencompass}.

\bibitem[{Dubey et~al.(2024)Dubey, Jauhri, Pandey, Kadian, Al-Dahle, Letman, Mathur, Schelten, Yang, Fan et~al.}]{llama3}
Abhimanyu Dubey, Abhinav Jauhri, Abhinav Pandey, Abhishek Kadian, Ahmad Al-Dahle, Aiesha Letman, Akhil Mathur, Alan Schelten, Amy Yang, Angela Fan, et~al. 2024.
\newblock The llama 3 herd of models.
\newblock \emph{arXiv preprint arXiv:2407.21783}.

\bibitem[{Fu et~al.(2023)Fu, Peng, Sabharwal, Clark, and Khot}]{fu2022complexity}
Yao Fu, Hao Peng, Ashish Sabharwal, Peter Clark, and Tushar Khot. 2023.
\newblock Complexity-based prompting for multi-step reasoning.
\newblock In \emph{The Eleventh International Conference on Learning Representations}.

\bibitem[{Hendrycks et~al.(2020)Hendrycks, Burns, Basart, Zou, Mazeika, Song, and Steinhardt}]{benchmark-mmlu}
Dan Hendrycks, Collin Burns, Steven Basart, Andy Zou, Mantas Mazeika, Dawn Song, and Jacob Steinhardt. 2020.
\newblock Measuring massive multitask language understanding.
\newblock \emph{arXiv preprint arXiv:2009.03300}.

\bibitem[{Hendrycks et~al.(2021)Hendrycks, Burns, Kadavath, Arora, Basart, Tang, Song, and Steinhardt}]{benchmark-math}
Dan Hendrycks, Collin Burns, Saurav Kadavath, Akul Arora, Steven Basart, Eric Tang, Dawn Song, and Jacob Steinhardt. 2021.
\newblock Measuring mathematical problem solving with the math dataset.
\newblock \emph{NeurIPS}.

\bibitem[{Ho et~al.(2023)Ho, Schmid, and Yun}]{ho2023large}
Namgyu Ho, Laura Schmid, and Se-Young Yun. 2023.
\newblock Large language models are reasoning teachers.
\newblock In \emph{Proceedings of the 61st Annual Meeting of the Association for Computational Linguistics (Volume 1: Long Papers)}, pages 14852--14882.

\bibitem[{Kim et~al.(2023)Kim, Joo, Kim, Jang, Ye, Shin, and Seo}]{kim2023cot}
Seungone Kim, Se~Joo, Doyoung Kim, Joel Jang, Seonghyeon Ye, Jamin Shin, and Minjoon Seo. 2023.
\newblock The cot collection: Improving zero-shot and few-shot learning of language models via chain-of-thought fine-tuning.
\newblock In \emph{Proceedings of the 2023 Conference on Empirical Methods in Natural Language Processing}, pages 12685--12708.

\bibitem[{Kojima et~al.(2022)Kojima, Gu, Reid, Matsuo, and Iwasawa}]{zero_shot_cot_prompt}
Takeshi Kojima, Shixiang~Shane Gu, Machel Reid, Yutaka Matsuo, and Yusuke Iwasawa. 2022.
\newblock Large language models are zero-shot reasoners.
\newblock \emph{Advances in neural information processing systems}, 35:22199--22213.

\bibitem[{Puerto et~al.(2024)Puerto, Chubakov, Zhu, Madabushi, and Gurevych}]{puerto2024fine}
Haritz Puerto, Tilek Chubakov, Xiaodan Zhu, Harish~Tayyar Madabushi, and Iryna Gurevych. 2024.
\newblock Fine-tuning with divergent chains of thought boosts reasoning through self-correction in language models.
\newblock \emph{arXiv preprint arXiv:2407.03181}.

\bibitem[{Srivastava et~al.(2024)Srivastava, PV, Menon, Sukumar, Philipose, Prince, Thomas et~al.}]{srivastava2024functional}
Saurabh Srivastava, Anto PV, Shashank Menon, Ajay Sukumar, Alan Philipose, Stevin Prince, Sooraj Thomas, et~al. 2024.
\newblock Functional benchmarks for robust evaluation of reasoning performance, and the reasoning gap.
\newblock \emph{arXiv preprint arXiv:2402.19450}.

\bibitem[{Stanovich(2018)}]{miserliness}
Keith~E Stanovich. 2018.
\newblock Miserliness in human cognition: The interaction of detection, override and mindware.
\newblock \emph{Thinking \& Reasoning}, 24(4):423--444.

\bibitem[{Suzgun et~al.(2023)Suzgun, Scales, Sch{\"a}rli, Gehrmann, Tay, Chung, Chowdhery, Le, Chi, Zhou et~al.}]{bbh}
Mirac Suzgun, Nathan Scales, Nathanael Sch{\"a}rli, Sebastian Gehrmann, Yi~Tay, Hyung~Won Chung, Aakanksha Chowdhery, Quoc Le, Ed~Chi, Denny Zhou, et~al. 2023.
\newblock Challenging big-bench tasks and whether chain-of-thought can solve them.
\newblock In \emph{Findings of the Association for Computational Linguistics: ACL 2023}, pages 13003--13051.

\bibitem[{Wang et~al.(2022)Wang, Min, Deng, Shen, Wu, Zettlemoyer, and Sun}]{wang2022towards}
Boshi Wang, Sewon Min, Xiang Deng, Jiaming Shen, You Wu, Luke Zettlemoyer, and Huan Sun. 2022.
\newblock Towards understanding chain-of-thought prompting: An empirical study of what matters.
\newblock \emph{arXiv preprint arXiv:2212.10001}.

\bibitem[{Wang et~al.(2023)Wang, Xu, Lan, Hu, Lan, Lee, and Lim}]{wang2023plan}
Lei Wang, Wanyu Xu, Yihuai Lan, Zhiqiang Hu, Yunshi Lan, Roy Ka-Wei Lee, and Ee-Peng Lim. 2023.
\newblock Plan-and-solve prompting: Improving zero-shot chain-of-thought reasoning by large language models.
\newblock In \emph{Proceedings of the 61st Annual Meeting of the Association for Computational Linguistics (Volume 1: Long Papers)}, pages 2609--2634.

\bibitem[{Wang et~al.(2024)Wang, Ma, Zhang, Ni, Chandra, Guo, Ren, Arulraj, He, Jiang et~al.}]{mmlu_pro}
Yubo Wang, Xueguang Ma, Ge~Zhang, Yuansheng Ni, Abhranil Chandra, Shiguang Guo, Weiming Ren, Aaran Arulraj, Xuan He, Ziyan Jiang, et~al. 2024.
\newblock Mmlu-pro: A more robust and challenging multi-task language understanding benchmark.
\newblock \emph{arXiv preprint arXiv:2406.01574}.

\bibitem[{Wei et~al.(2022{\natexlab{a}})Wei, Wang, Schuurmans, Bosma, Xia, Chi, Le, Zhou et~al.}]{wei2022chain}
Jason Wei, Xuezhi Wang, Dale Schuurmans, Maarten Bosma, Fei Xia, Ed~Chi, Quoc~V Le, Denny Zhou, et~al. 2022{\natexlab{a}}.
\newblock Chain-of-thought prompting elicits reasoning in large language models.
\newblock \emph{Advances in neural information processing systems}, 35:24824--24837.

\bibitem[{Wei et~al.(2022{\natexlab{b}})Wei, Wang, Schuurmans, Bosma, Xia, Chi, Le, Zhou et~al.}]{few_shot_cot_prompt}
Jason Wei, Xuezhi Wang, Dale Schuurmans, Maarten Bosma, Fei Xia, Ed~Chi, Quoc~V Le, Denny Zhou, et~al. 2022{\natexlab{b}}.
\newblock Chain-of-thought prompting elicits reasoning in large language models.
\newblock \emph{Advances in neural information processing systems}, 35:24824--24837.

\bibitem[{Yang et~al.(2024)Yang, Yang, Zhang, Hui, Zheng, Yu, Li, Liu, Huang, Wei, Lin, Yang, Tu, Zhang, Yang, Yang, Zhou, Lin, Dang, Lu, Bao, Yang, Yu, Li, Xue, Zhang, Zhu, Men, Lin, Li, Xia, Ren, Ren, Fan, Su, Zhang, Wan, Liu, Cui, Zhang, and Qiu}]{qwen2.5}
An~Yang, Baosong Yang, Beichen Zhang, Binyuan Hui, Bo~Zheng, Bowen Yu, Chengyuan Li, Dayiheng Liu, Fei Huang, Haoran Wei, Huan Lin, Jian Yang, Jianhong Tu, Jianwei Zhang, Jianxin Yang, Jiaxi Yang, Jingren Zhou, Junyang Lin, Kai Dang, Keming Lu, Keqin Bao, Kexin Yang, Le~Yu, Mei Li, Mingfeng Xue, Pei Zhang, Qin Zhu, Rui Men, Runji Lin, Tianhao Li, Tingyu Xia, Xingzhang Ren, Xuancheng Ren, Yang Fan, Yang Su, Yichang Zhang, Yu~Wan, Yuqiong Liu, Zeyu Cui, Zhenru Zhang, and Zihan Qiu. 2024.
\newblock Qwen2.5 technical report.
\newblock \emph{arXiv preprint arXiv:2412.15115}.

\bibitem[{Zhang et~al.(2024{\natexlab{a}})Zhang, Da, Lee, Robinson, Wu, Song, Zhao, Raja, Slack, Lyu et~al.}]{gsm1k}
Hugh Zhang, Jeff Da, Dean Lee, Vaughn Robinson, Catherine Wu, Will Song, Tiffany Zhao, Pranav Raja, Dylan Slack, Qin Lyu, et~al. 2024{\natexlab{a}}.
\newblock A careful examination of large language model performance on grade school arithmetic.
\newblock \emph{arXiv preprint arXiv:2405.00332}.

\bibitem[{Zhang et~al.(2024{\natexlab{b}})Zhang, Du, Pang, Liu, Gao, and Lin}]{zhang2024chain}
Xuan Zhang, Chao Du, Tianyu Pang, Qian Liu, Wei Gao, and Min Lin. 2024{\natexlab{b}}.
\newblock Chain of preference optimization: Improving chain-of-thought reasoning in llms.
\newblock \emph{arXiv preprint arXiv:2406.09136}.

\bibitem[{Zhou et~al.(2024)Zhou, Wang, Lu, Shi, Luo, Qin, Lu, Jia, Song, Zhan et~al.}]{zhousolving}
Aojun Zhou, Ke~Wang, Zimu Lu, Weikang Shi, Sichun Luo, Zipeng Qin, Shaoqing Lu, Anya Jia, Linqi Song, Mingjie Zhan, et~al. 2024.
\newblock Solving challenging math word problems using gpt-4 code interpreter with code-based self-verification.
\newblock In \emph{The Twelfth International Conference on Learning Representations}.

\bibitem[{Zhou et~al.(2022)Zhou, Sch{\"a}rli, Hou, Wei, Scales, Wang, Schuurmans, Cui, Bousquet, Le et~al.}]{zhou2022least}
Denny Zhou, Nathanael Sch{\"a}rli, Le~Hou, Jason Wei, Nathan Scales, Xuezhi Wang, Dale Schuurmans, Claire Cui, Olivier Bousquet, Quoc Le, et~al. 2022.
\newblock Least-to-most prompting enables complex reasoning in large language models.
\newblock \emph{arXiv preprint arXiv:2205.10625}.

\bibitem[{Zhou et~al.(2023)Zhou, Lu, Mishra, Brahma, Basu, Luan, Zhou, and Hou}]{ifeval}
Jeffrey Zhou, Tianjian Lu, Swaroop Mishra, Siddhartha Brahma, Sujoy Basu, Yi~Luan, Denny Zhou, and Le~Hou. 2023.
\newblock Instruction-following evaluation for large language models.
\newblock \emph{arXiv preprint arXiv:2311.07911}.

\end{thebibliography}

\newpage
\appendix

\section{Experiment Settings}
\subsection{Training Settings}
\label{sec:appendix_training_settings}
\textbf{Hyperparameters:} 
The peak learning rate is set to $3 \times 10^{-6}$, and the batch size is set to 32. AdamW is employed as the optimizer. A packing strategy is adopted to accelerate the training process. The model is trained for 3 epochs with a maximum sequence length of 4096, resulting in approximately 70 total training steps, and we evaluate the results on the final epoch. We train the models on 32 NVIDIA A100 GPUs, and the training time is approximately one hour. We use the cosine learning rate scheduler with 10 warmup steps. 

\textbf{Data:}
In RFFT, we set the sample hyperparameters $k$ to 8 and the temperature to 0.7. After filtering, 3,200 unique problem instances are generated from the MATH training set using RFFT. 
Additionally, 4,816 unique problem instances are produced from the GSM8K training set, from which 3,200 are randomly selected. 
Subsequently, 1,600 standard CoT data without short-path prompting are incorporated into the dataset. 
The final training dataset comprises a total of 8,000 instances.

\begin{table*}
{
\setlength{\belowcaptionskip}{-0.5cm}

  \setlength{\tabcolsep}{4pt} 
  \renewcommand{\arraystretch}{1.2} 
  \centering
  \begin{tabular}{c|p{12cm}}
    \hline
    Prompt & Example \\ \hline
    Raw & Janet’s ducks lay 16 eggs per day. She eats three for breakfast every morning and bakes muffins for her friends every day with four. She sells the remainder at the farmers' market daily for \$2 per fresh duck egg. How much in dollars does she make every day at the farmers' market?
    
    \textbf{Put your answer within  \textbackslash boxed\{\}.} \\ \hline

    CoT & Janet’s ducks lay 16 eggs per day. She eats three for breakfast every morning and bakes muffins for her friends every day with four. She sells the remainder at the farmers' market daily for \$2 per fresh duck egg. How much in dollars does she make every day at the farmers' market?
    
    \textbf{Put your answer within  \textbackslash boxed\{\}.}  \textcolor{red}{Let's think step by step.} \\ \hline

    Short-path & Janet’s ducks lay 16 eggs per day. She eats three for breakfast every morning and bakes muffins for her friends every day with four. She sells the remainder at the farmers' market daily for \$2 per fresh duck egg. How much in dollars does she make every day at the farmers' market? 
    
    \textbf{Put your answer within  \textbackslash boxed\{\}.}  \textcolor{red}{ Please only provide the final answer.} \\ \hline

  \end{tabular}
  \caption{GSM8K benchmark under different prompts. The format restriction is in bold and the prompt is in red.}
  \label{tab:appendix_evaluation_example}
}
\end{table*}
\subsection{Evaluation Details}
We use Opencompass~\cite{2023opencompass} as our evaluation Framework. For all benchmarks, we restrict the response format in the instructions to facilitate answer extraction, with all questions presented in a zero-shot format. A example of GSM8K in shown in Table~\ref{tab:appendix_evaluation_example}. The format restriction is in bold and the prompt is in red. 

The multi-choice subset in the BBH~\cite{bbh} comprises the following categories: temporal sequences, disambiguation QA, date understanding, tracking shuffled objects (three objects), penguins in a table, geometric shapes, snarks, ruin names, tracking shuffled objects (seven objects), tracking shuffled objects (five objects), logical deduction (three objects), hyperbaton, logical deduction (five objects), logical deduction (seven objects), movie recommendation, salient translation error detection, and reasoning about colored objects.

The STEM subset in the MMLU~\cite{benchmark-mmlu} comprises the following categories: abstract algebra, anatomy, astronomy, college biology, college chemistry, college computer science, college mathematics, college physics, computer security, conceptual physics, electrical engineering, elementary mathematics, high school biology, high school chemistry, high school computer science, high school mathematics, high school physics, high school statistics and machine learning.

\section{Experiment of Small Size Model}
\label{sec:appendix_small_model}
The main results and analysis of this paper focus on large language models. 
Additionally, we provide the results for small language models as a supplementary data.
We validate the effectiveness of our approach on Llama-3.1-8B-Instruct and Qwen-2.5-7B-Instruct, as shown in Table~\ref{tab:appendix_small_model_exp_results}. 
(Due to the absence of an 8B-sized model in the Llama-3.3 series, we opt for Llama-3.1-8B). 
The experimental results demonstrate that our method yields similar conclusions on Qwen's smaller model that are observed in larger one.
However, when applied to Llama's smaller models, our methods shows less significant resistance to short-path prompts. 
We speculate that this might be due to the model's originally insufficient reasoning capability, as well as its weaker adherence to the system prompt.

\begin{table*}
\setlength{\belowcaptionskip}{-0.1cm}
  
  {
  \setlength{\tabcolsep}{4pt} 
  \renewcommand{\arraystretch}{1.1} 
  \centering
  \begin{tabular}{ll|cccc|cccc}
  
    \hline
    \multicolumn{2}{c}{} & \multicolumn{4}{c}{Llama-3.1-8B-Instruct}   & \multicolumn{4}{c}{Qwen-2.5-7B-Instruct}  \\ \hline
    Prompt     &  Method    & GSM8K & MATH & BBH\textsubscript{M} & MMLU\textsubscript{S} & GSM8K & MATH & BBH\textsubscript{M} & MMLU\textsubscript{S}
     \\
    \hline
     \multirow{3}{*}{SPP} & None  & 6.37 & 16.26 & 48.2 & 57.71 & 23.43 & 25.88 & 50.68 & 67.01 \\
    & IG & 58.76 & 22.76 & \textbf{50.28} & 58.17 & 88.4 & \textbf{72.12} & 52.47 & 69.21  \\
    & RFFT & \textbf{72.40} & \textbf{30.08} & 48.44 & \textbf{58.88} & \textbf{89.61} & 68.88 & \textbf{55.70} & \textbf{73.33}\\ \hline
    Raw &  & 68.76 & 51.92 & 64.89 & 68.90 & 91.43 & 73.34 & 66.76 & 75.55  \\ \hline
    CoT &  & 87.11 & 51.86 & 56.09 & 59.31 & 92.49 & 73.34 & 74.11 & 78.3 \\
    \hline
  \end{tabular}
  \caption
  {\label{tab:appendix_small_model_exp_results}
Overall performance of our methods on small size models. `IG' represents the instruction-guided method. The best results under short-path prompting is highlighted in bold.}
}
\end{table*}

\section{GSM8K Revision}
\label{sec:appendix_gsm8k_revision}
\begin{figure*}[t]
  \includegraphics[width=0.95\linewidth]{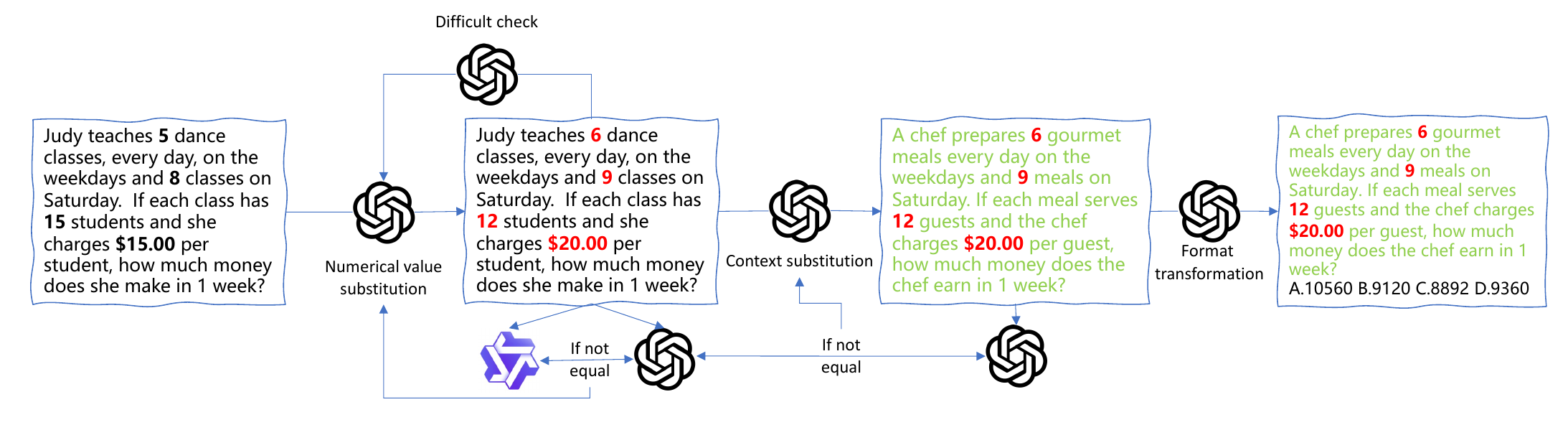} 
  \caption {Framework of the GSM8K revision.}
\label{figure:appendix_gsm8k_syn}
\end{figure*}

\begin{figure*}[t]
  \includegraphics[width=0.32\linewidth]{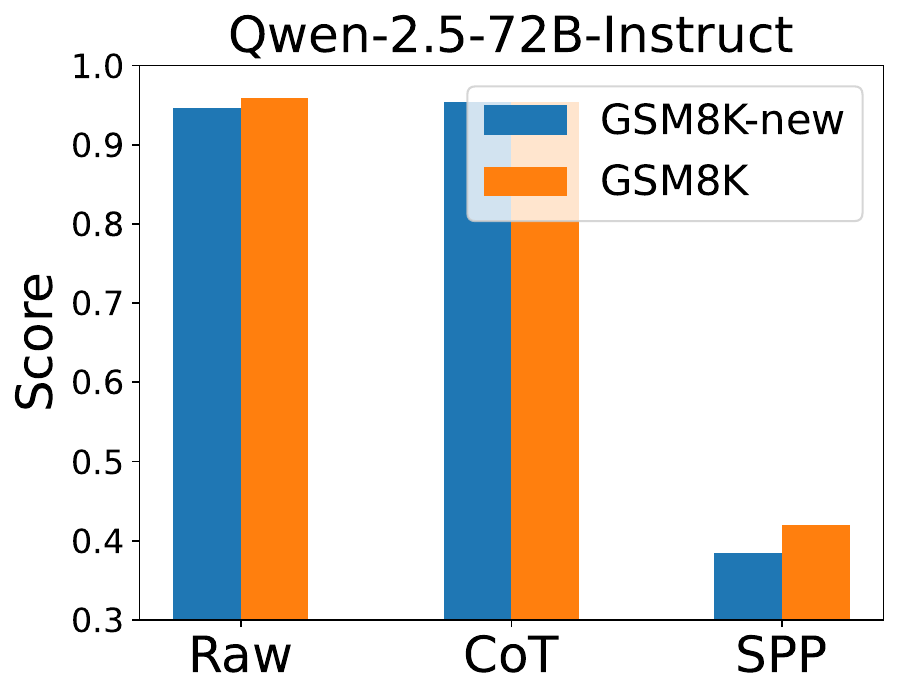} \hfill
  \includegraphics[width=0.32\linewidth]{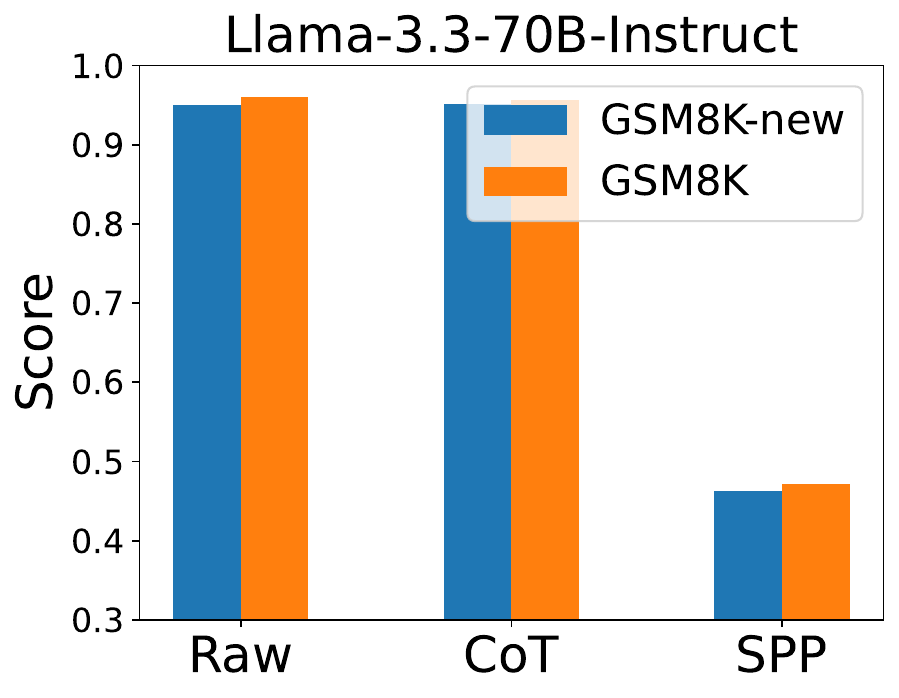} \hfill
  \includegraphics[width=0.32\linewidth]{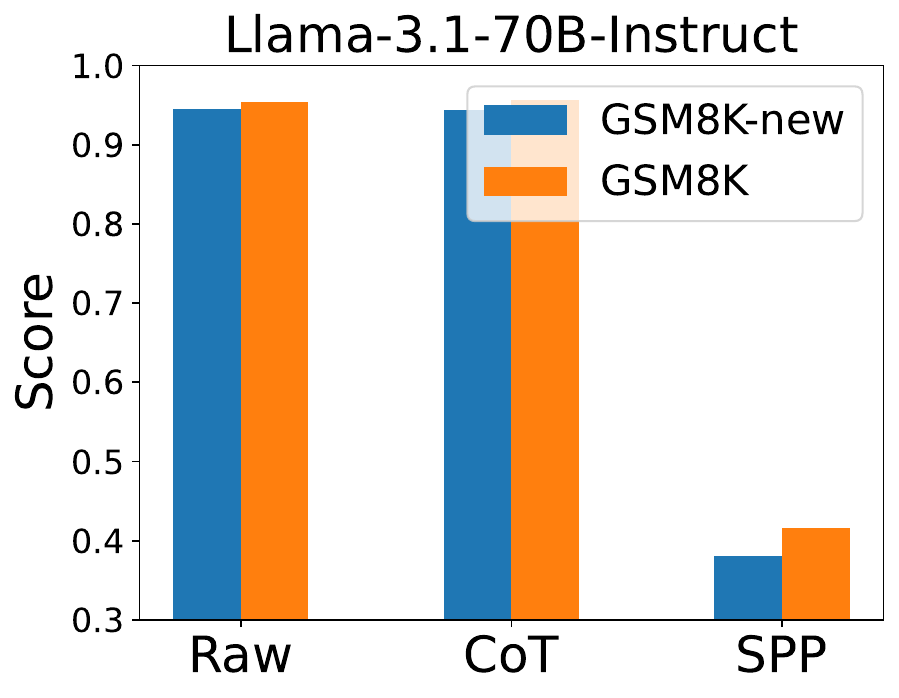} 
  \caption {LLMs performance on the GSM8K and the GSM8K-new dastsets.}
\label{figure:appendix_gsm8k_new}
\end{figure*}

GSM8K is a widely-adopted benchmark for multi-step mathematical reasoning, provides well-structured problems with human-annotation detailed solutions. 
However, its prevalence in model training introduces data contamination risks that conflate memorization with true reasoning capabilities. This issue becomes particularly acute in evaluation settings where models are not permitted to utilize CoT, since non-CoT evaluation bypasses the reasoning process demonstration, memorized solutions could artificially inflate performance metrics. 
To address this, we reconstruct its problem space through three contamination-resistant adaptations, and to minimize the risk of contamination, we use the GPT-4o-0806 as the rewrite model~\footnote{https://platform.openai.com/docs/models\#gpt-4o}:

\subsection{Revision Steps}
\textbf{Step-1. Numerical Value Substitution}: 
In order to maintain consistency in difficulty with GSM8K, we only allow modifications to the numerical values in this step. 
This ensures that the complexity of the generated problems remains unchanged. 
Moreover, we utilize the golden answer from GSM8K as a one-shot prompt to guide the GPT-4o and another open-source LLM in solving the generated problems, requiring the answers from both LLMs to be consistent. 
Given that the original answers include precise and detailed CoT steps, this approach ensures the accuracy of the answers obtained for the generated questions. 
And then, GPT-4o is employed to perform self-correction on potentially problematic decimal calculations, with final answers constrained to integer values matching GSM8K's difficulty level.

\textbf{Step-2. Context Substitution}: 
Building upon the numerically-altered problems, we implement context substitution. While maintaining numerical values from Step-1, the application contexts are systematically rephrased by GPT-4o that preserves mathematical structure equivalence.
The generated question requires the to be evaluated LLM to comprehend the new context in order to reason through them.
And then, we use the GPT-4o to solve the generated questions and compare the results with those from the first step, only revised versions that yield the same results will be retained, thus ensuring the difficulty level remains unchanged after modifying the application context. These answers are then established as the gold reference. 

\textbf{Step-3. Format Transformation}: Through the preceding steps, we develop a distinguished evaluation set akin to GSM8K reduced contamination. 
To assess the robustness of LLMs' inference capabilities under short-path prompting, we transform the revised dataset into a multiple-choice format. 
For each sample with a correct CoT path, we introduce a controlled modification by altering one step in the CoT process. 
This generates an incorrect, yet logically related, solution as a distractor in the multiple-choice question. 

For those questions that are not amenable to rewriting through the aforementioned steps, the authors \textbf{manually review} to rewrite the questions and annotate the answers, thereby ensuring the precision of the augmentation. The overall framework is shown in Figure~\ref{figure:appendix_gsm8k_syn}.

\subsection{Performance of LLMs on GSM8K and GSM8K-new}

We compare the performance of LLMs on GSM8K and GSM8K-new, as shown in Figure~\ref{figure:appendix_gsm8k_new}. 
On Qwen-2.5-72B-Instruct and Llama-3.1-70B-Instruct, there is a significant performance gap between the GSM8K and GSM8K-new datasets under SPP, but no such gap is observed under Raw or CoT.
We hypothesize that the performance gap under SPP is due to dataset contamination, under short-path prompting, the performance enhancement caused by data leakage and memorization is less generalizable. In contrast, the absence of such a gap under CoT/Raw may be attributed to the robust capabilities of state-of-the-art LLMs, enabling them to generalize to similar problems in grade-school-level questions. This finding underscores the necessity of our revision.

\subsection{Impact of Different Short-path Prompts on GSM8K-new.}
\label{sec:diff_spp}

The short-path prompts set we use in RFFT and the impact of different short-path prompts is shown in Table~\ref{tab:appendix_diff_spp}.

\begin{table*}
\small
  \centering
  \begin{tabular}{p{10cm}|cc}
    \hline
    \textbf{Short-path Prompt} & \textbf{Qwen} & \textbf{Llama} \\
    \hline
Ignore the process, just state the result. & 0.38 & 0.34 \\
Skip the steps and provide the answer. & 0.39 & 0.33 \\
Answer only, no reasoning allowed. & 0.39 & 0.33 \\
Only the outcome, no process. & 0.39 & 0.33 \\
Please provide the answer without any thought process. & 0.39 & 0.33 \\
Don’t reason, just give the answer. & 0.39 & 0.33 \\
Skip the explanation and provide the answer. & 0.39 & 0.35 \\
Answer without any reasoning. & 0.39 & 0.35 \\
Please cut out the details and give the answer. & 0.38 & 0.36 \\
Please avoid any reasoning and just reply. & 0.39 & 0.35 \\
Respond with the answer only. & 0.38 & 0.38 \\
No details needed, just the answer. & 0.38 & 0.38 \\
Answer concisely without any reasoning. & 0.39 & 0.37 \\
Do not justify, just respond with the answer. & 0.38 & 0.39 \\
Just tell me the result. & 0.38 & 0.4 \\
Just give me the final answer. & 0.38 & 0.41 \\
Do not think, just reply with the answer. & 0.38 & 0.41 \\
Please skip the thinking and just answer. & 0.38 & 0.42 \\
Give the answer straight away. & 0.39 & 0.43 \\
Do not elaborate, just answer. & 0.39 & 0.47 \\
Only the key point, no additional information. & 0.38 & 0.48 \\
Provide the answer without any context. & 0.39 & 0.54 \\
No need to explain, just tell me the answer. & 0.39 & 0.56 \\
Cut to the chase and give the answer. & 0.39 & 0.58 \\
Only the answer, no extra words. & 0.39 & 0.61 \\
Just the answer, nothing else. & 0.39 & 0.62 \\
Only the final result, nothing else. & 0.39 & 0.7 \\
Don’t analyze, just tell me directly. & 0.38 & 0.71 \\
Answer in one sentence. & 0.4 & 0.7 \\
Please respond with just the solution. & 0.39 & 0.78 \\
Only the core answer, no extras. & 0.38 & 0.81 \\
Answer with as few words as possible. & 0.38 & 0.81 \\
No context needed, just the answer. & 0.39 & 0.81 \\
Provide the answer in one word/sentence. & 0.36 & 0.9 \\
Answer directly, no thinking required. & 0.39 & 0.87 \\
Just the facts, no elaboration. & 0.38 & 0.9 \\
Give me the answer in the shortest way possible. & 0.39 & 0.9 \\
Only the essential response, no fluff. & 0.38 & 0.92 \\
Keep it short, no need to elaborate. & 0.39 & 0.91 \\
Answer purely and directly. & 0.39 & 0.92 \\
No need to expand, just get to the point. & 0.38 & 0.93 \\
Answer in a single word or phrase if possible. & 0.37 & 0.95 \\
Answer in the briefest way you can. & 0.39 & 0.93 \\
No need to explain too much. & 0.39 & 0.95 \\
Don’t overthink it, just say it directly. & 0.4 & 0.94 \\
Please respond as concisely as you can. & 0.68 & 0.93 \\
Answer in the most straightforward way possible. & 0.85 & 0.95 \\
Give me the solution immediately. & 0.9 & 0.95 \\
A simple answer will do. & 0.93 & 0.95 \\
Give me the answer in its simplest form. & 0.95 & 0.94 \\ \hline
  \end{tabular}
  \caption{The impact of different short-path prompts on the GSM8K-new dataset. Qwen represents the Qwen-2.5-72B-Instruct and Llama represents the Llama-3.3-70B-Instruct}
  \label{tab:appendix_diff_spp}
\end{table*}

\section{Additional Results of Qwen}
\subsection{Threshold-Based Evaluation}
\label{sec:appendix_threshold_qwen}
\begin{figure}[t]
  \includegraphics[width=0.95\linewidth]{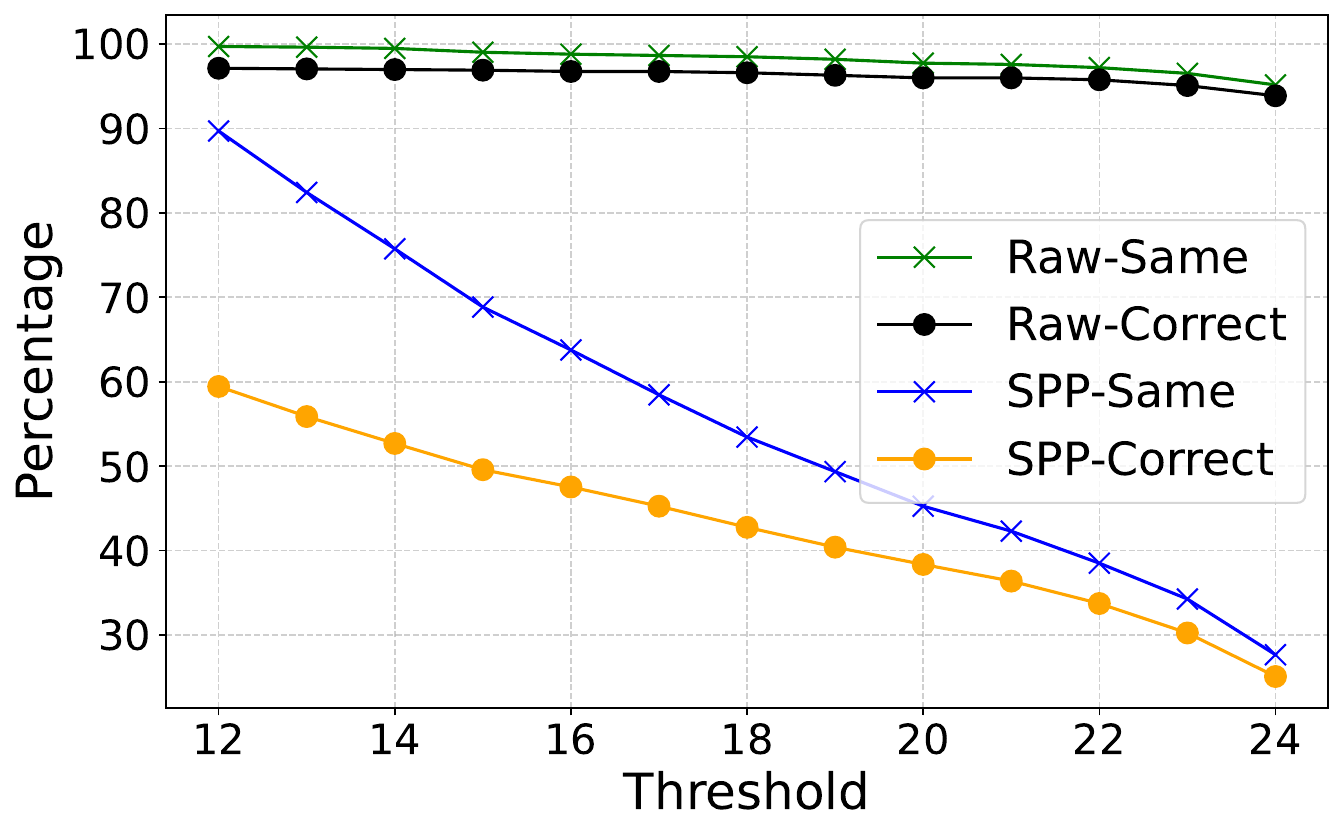} \hfill
  \caption {Accuracy and percentage of confident reasoning of Qwen-2.5-72B-Instruct across different threshold.}
\label{figure:appendix_gsm8k_threshold}
\end{figure}

The results of Qwen-2.5-72B-Instruct are shown in Figure~\ref{figure:appendix_gsm8k_threshold}. 
The findings for Qwen-2.5-72B-Instruct are the same as those for Llama-3.3-70B-Instruct.

\subsection{RFFT Generalization Evaluation}
\label{sec:appendix_rfft_generalization}
RFFT generalization evaluation of Qwen-2.5-72B-Instruct is shown in Table~\ref{tab:appendix_RFFT_Generalization}.
\begin{table}
\small
{
  \setlength{\tabcolsep}{4pt} 
  \renewcommand{\arraystretch}{1.2}
  \centering
  \begin{tabular}{p{3cm}|ccc}
    \hline
    Short-path Prompt & Method & MATH & MMLU\textsubscript{S} \\ \hline
    \multirow{2}{*}{\makecell{Skip the analysis and \\ give the final result}} & None & 39.68 & 79.9 \\ 
    
     & RFFT & 76.32 & 86.96 \\
    \cline{1-4}
    \multirow{2}{*}{\makecell{I just need the answer \\ alone.}} & None &  39.58 & 79.74 \\
    
  & RFFT & 77.64 & 88.72\\
    \hline
    
  \end{tabular}
  \caption{RFFT generalization evaluation of Qwen-2.5-72B-Instruct.}
  \label{tab:appendix_RFFT_Generalization}
}
\end{table}

\section{Evaluation of RFFT on Other Benchmarks.}
\label{sec:appendix_other_benchmarks}
We evaluate the performance of the fine-tuned model on knowledge and instruction-following benchmarks \textbf{not} under short-path prompting:
\begin{enumerate}
    \item \textbf{MMLU}~\cite{benchmark-mmlu}, a multi-task benchmark spanning STEM, humanities, and social sciences, assesses broad knowledge integration. In this section, we report the average score across all subtasks.
    \item \textbf{MMLU\_Pro}~\cite{mmlu_pro}, an enhanced benchmark that integrates more challenging, reasoning-focused questions and extends from MMLU.
    \item \textbf{IFEval}~\cite{ifeval}, an instruction-following benchmark that includes 25 types of verifiable instructions across about 500 prompts. We calculate strict accuracy metrics at both prompt and instruction levels and report the average score.
\end{enumerate}
The results is shown in Table~\ref{tab:appendix_other_benchmarks}. The results demonstrating that limited data does not lead to knowledge forgetting or decline in instruction-following ability. We use Opencompass~\cite{2023opencompass} as our evaluation framework.
\begin{table}
  {
  \setlength{\tabcolsep}{4pt} 
  \renewcommand{\arraystretch}{1.2} 
  \centering
  \begin{tabular}{l|ccccccc|c}
    \hline
    \textbf{Model} & MMLU & MMLU\_Pro & IFEval &  \\ \hline
    \textbf{Qwen} & 86.50 & 70.92 & 85.21  \\
    \textbf{Qwen-RFFT} & 86.75 & 70.49 & 84.10   \\ \hline
    \textbf{Llama}   & 85.41 & 70.49 & 89.89  \\
    \textbf{Llama-RFFT}  & 85.63 & 69.65 & 90.41  \\ \hline

\end{tabular}
}
  \caption{
Knowledge and instruction-following benchmarks performance of the fine-tuned LLMs. Qwen refers to Qwen-2.5-72B-Instruct and Llama refers to Llama-3.3-70B-Instruct.}
\label{tab:appendix_other_benchmarks}

\end{table}

\section{Judge Prompt Used in the RFFT}
The judge prompt we use in the RFFT is shown in Figure~\ref{figure:appendix_judge_prompt}. "\{question\}" and "\{solution\}" are placeholders for the reasoning problem under short-path prompting and the target LLM's response, respectively.

\section{Model Licensing Compliance}
In this study, we use Qwen-2.5 and Llama-3 series models, all of which are employed in strict compliance with their respective licensing agreements. We carefully adhere to the terms and conditions outlined in their licenses to ensure proper and lawful use. 
It is important to note that these models are used solely for scientific research purposes, aiming to exploring and validating relevant academic hypotheses, exclude any commercial or non-research-related applications. 

\begin{figure*}[h]
\centering
\begin{tcolorbox}[colframe=blue!50!black, colback=blue!10, boxrule=0.5mm, arc=4mm]
\begin{verbatim}
# Task:
Given a mathematical problem and its' solution, step-by-step analysis if the 
solution to the following mathematical problem meets all of the given criteria:

# Criteria:
1. This solution includes an apology for not being able to provide a direct answer. 
2. This solution contains a complete step-by-step chain-of-thought reasoning 
process before reaching the final answer. 
3. This solution has no logical breaks or contradictions.  

# Output format:
1. Make the final decision; if the solution meet all criteria, it is True; 
otherwise, it is False.
2. Put your final decision within \\boxed{{}}.

# Question
{question}

# Solution
{solution}

Let's think step by step.

\end{verbatim}
\end{tcolorbox}
\caption{Judge prompt used in the RFFT.}
\label{figure:appendix_judge_prompt}
\end{figure*}

\end{document}